\documentclass[letterpaper, 10 pt, conference]{ieeeconf}  

\IEEEoverridecommandlockouts                         
\usepackage{graphics} 
\usepackage{mathptmx} 
\usepackage{amsmath} 
\usepackage{amssymb}  

\usepackage[noadjust]{cite} 
\usepackage[capitalize]{cleveref} 

\crefname{section}{Sec.}{Sec.}

\usepackage{siunitx}
\usepackage{scalerel} 
\usepackage{mathtools} 
\newcommand{\minus}{\scalebox{0.75}[1.0]{$-$}} 
\newcommand{\mminus}{\scalebox{0.65}[1.0]{$-$}} 
\usepackage{xfrac} 
\newcommand{\mytilde}{\raise.17ex\hbox{$\scriptstyle\mathtt{\sim}$}} 

\usepackage{booktabs} 

\usepackage{adjustbox} 
\usepackage{nccmath}  
\usepackage{floatrow} 
\usepackage{subcaption} 
\usepackage{dblfloatfix}
\usepackage{afterpage}
\usepackage{xcolor}
\usepackage{xspace} 

\DeclareCaptionLabelSeparator{periodspace}{.\quad}
\usepackage{tikz}
\usetikzlibrary{shapes, calc}

\newcommand{\VP}{\textrm{VP}\xspace}

\newcommand{\GRF}{\textrm{GRF}\xspace}
\newcommand{\CoM}{\textrm{CoM}\xspace}
\newcommand{\TSLIP}{\textrm{TSLIP}\xspace}
\newcommand{\GRFx}{$\mathrm{GRF_{x}}$\xspace}
\newcommand{\GRFy}{$\mathrm{GRF_{y}}$\xspace}
\newcommand{\VPa}{$\mathrm{VP_{\color{color_VPBlue}{A}}}$\xspace}
\newcommand{\VPb}{$\mathrm{VP_{\color{color_VPRose}{B}}}$\xspace} 
\newcommand{\VPbl}{$\mathrm{VP_{\color{color_VPRose}{BL}}}$\xspace} 

\newcommand{\markerCW}{\raisebox{-1 pt}{\tikz{\node[draw=white,scale=0.6,circle,fill={color_CW},opacity=.5](){};}}}
\newcommand{\markerCCW}{\raisebox{-1 pt}{\tikz{\node[draw=white,scale=0.6,circle,fill={color_CCW}](){};}}}

\newcommand{\markerVPa}{\raisebox{-1 pt}{\tikz{\node[draw=white,scale=0.6,circle,fill={color_VPBlue}](){};}}}
\newcommand{\markerVPb}{\raisebox{-1 pt}{\tikz{\node[draw=white,scale=0.6,circle,fill={color_VPRose}](){};}}}
\newcommand{\markerVPbA}{\raisebox{0 pt}{\tikz{\node[draw=white,scale=0.7,shape=rectangle,fill={color_VPb_A},opacity=.9](){};}}\xspace}
\newcommand{\markerVPbB}{\raisebox{0 pt}{\tikz{\node[draw=white,scale=0.7,shape=rectangle,fill={color_VPb_B},opacity=.9](){};}}\xspace}

\newcommand{\arrowWorkP}{\raisebox{-2 pt}{\tikz{\draw[->, dash pattern={on 3pt off 1 pt on 1pt off 1pt on 3pt off 1 pt on 1pt off 1pt}, line width=0.35mm, {color_Wp}](0, 0.01)--(0, 0.35);}}}
\newcommand{\arrowWorkN}{\raisebox{-2 pt}{\tikz{\draw[->, dash pattern={on 3pt off 1 pt on 1pt off 1pt on 3pt off 1 pt on 1pt off 1pt}, line width=0.35mm, {color_Wn}](0, 0.35)--(0, 0.01);}}}

 \newcommand{\markerVPaTri}{\raisebox{-2 pt}{\tikz{\node[draw=white, isosceles triangle, isosceles triangle apex angle=60, inner sep=0pt, anchor=lower side, rotate=0, line width=1.3pt, minimum height=0.24cm, fill={color_VPBlue}] (triangle) at (0,0) {};}}\!\xspace}
 \newcommand{\markerVPbTri}{\raisebox{-2 pt}{\tikz{\node[draw=white, isosceles triangle, isosceles triangle apex angle=60, inner sep=0pt, anchor=lower side, rotate=0, line width=1.3pt, minimum height=0.24cm, fill={color_VPRose}] (triangle) at (0,0) {};}}\!\xspace}

\newcommand{\markerHuMoD}{\raisebox{-1 pt}{\tikz{\node[draw=white,scale=0.6,circle,fill={color_HuMod}](){};}}}

 \newcommand{\markerDmpLin}{\raisebox{-1 pt}{\tikz{\node[draw=white, isosceles triangle, isosceles triangle apex angle=60, inner sep=0pt, anchor=lower side, rotate=90, line width=1.5pt, minimum height=0.27cm, fill={color_Linear}] (triangle) at (0,0) {};}}\xspace}

\newcommand{\markerDmpBiLin}{\raisebox{0 pt}{\tikz{\node[draw=white,scale=0.7,shape=rectangle,fill={color_Biliniar}](){};}}\xspace}

\title{\LARGE \bf
Trunk Pitch Oscillations for Joint Load Redistribution \\ in Humans and Humanoid Robots
}

\author{{\"O}zge Drama and Alexander Badri-Spr{\"o}witz
\thanks{The authors thank the International Max Planck Research School for Intelligent Systems (IMPRS-IS) for supporting {\"O}zge Drama.}
\thanks{{\"O}zge Drama and Alexander Badri-Spr{\"o}witz are with the Dynamic Locomotion Group at Max Planck Institute for Intelligent Systems, Stuttgart, Germany {\tt\small \{drama,sprowitz\}@is.mpg.de}}%
}

\begin{document}

\definecolor{color_VPRose}{rgb}{0.6980    0.0941    0.1686} 
\definecolor{color_VPBlue}{rgb}{0.1294    0.4000    0.6745}
\definecolor{color_VPb_B}{rgb}{ 0.8978    0.4922    0.5862}
\definecolor{color_VPb_A}{rgb}{0.9488    0.7460    0.6982}

\definecolor{color_CW}{rgb}{0.5020    0.8196    0.8510}
\definecolor{color_CCW}{rgb}{0.9922    0.6314    0.4392}
\definecolor{color_PowderBlue}{rgb}{0.7176    0.8706    0.8431}
\definecolor{color_Thistle}{rgb}{0.8078    0.7373    0.8667}
 \definecolor{color_Seagreen}{rgb}{0.3528    0.5041    0.4233}
 \definecolor{color_BlueGrey}{rgb}{0.4 0.4 0.4}
\definecolor{color_HuMod}{rgb}{0.5020    0.8039    0.7569}
\definecolor{color_Biliniar}{rgb}{0.1686    0.5490    0.7451}
\definecolor{color_Linear}{rgb}{0.0314    0.2510    0.5059}
\definecolor{color_Wp}{rgb}{0.584 0.643 0.737} 
\definecolor{color_Wn}{rgb}{0.666  0.596 0.611} 

\maketitle
\thispagestyle{empty}
\pagestyle{empty}

\begin{abstract} %
Creating natural-looking running gaits for humanoid robots is a complex task due to the underactuated degree of freedom in the trunk, which makes the motion planning and control difficult.
The research on trunk movements in human locomotion is insufficient, and no formalism is known to transfer human motion patterns onto robots. Related work mostly focuses on the lower extremities, and simplifies the problem by stabilizing the trunk at a fixed angle.
In contrast, humans display significant trunk motions that follow the natural dynamics of the gait.
In this work, we use a spring-loaded inverted pendulum model with a trunk (\TSLIP) together with a virtual point (\VP) target to create trunk oscillations and investigate the impact of these movements. 
We analyze how the \VP location and forward speed determine the direction and magnitude of the trunk oscillations. We show that positioning the \VP below the center of mass (\CoM) can explain the forward trunk pitching observed in human running. The \VP below the \CoM leads to a synergistic work between the hip and leg, reducing the leg loading. However, it comes at the cost of increased peak hip torque.
Our results provide insights for leveraging the trunk motion to redistribute joint loads and potentially improve the energy efficiency in humanoid robots.
\end{abstract}

\section{INTRODUCTION} \label{sec:Intro} 

Traditional humanoid research considers trunk movements as \emph{undesired} and designs controllers to keep the trunk at a fixed angle \cite{Raibert_1986, Westervelt_2007}, which leads to \emph{stiff and fixed} upper bodies. Fixed trunk postures are in sharp contrast with the human locomotion, that exhibits pronounced trunk oscillations of up to $\pm$ \SIrange[range-phrase=-,range-units=single]{2}{6} in the sagittal plane \cite{Aminiaghdam_2017, Heitcamp_2012, Hinrichs_1987, Schache_1999, Thorstensson_1984}. The trunk matters in bipedal anatomy, as it is inherently unstable, comprises \SI{50}{\percent} of total body mass, and has a large inertia \cite{deLeva_1996, Thorstensson_1984}. Thus, even small deviations in trunk position can significantly effect the whole body dynamics \cite{Aminiaghdam_2017}. In this paper, we use a spring-loaded inverted pendulum model with a controller based on a virtual point (\VP) to generate trunk pitch oscillations with varying magnitudes and direction. We investigate how the trunk angular excursion and the direction of the oscillation depend on the \VP position and the running speed. 

Humans utilize the trunk's inertia extensively to assist locomotion; especially during running to offset the highly accelerated swing motion and counteract destabilizing, high ground reaction forces (\GRF) \cite{Bramble_2004}.  In running, the human trunk pitches forward (anteriorly) during the first half of the stance phase, reaches a maximum flexion at mid-stance and pitches backward (posteriorly) in the second half of the stance phase \cite{Heitcamp_2012, Hinrichs_1987, Thorstensson_1984}. In walking, the trunk moves forward in the double stance phase. These oscillations can be characterized by four parameters (i)-(iv) \cite{Heitcamp_2012}, which are adjusted in response to variations in speed, mode of progression (e.g., gait type, motion direction) and terrain conditions. 

\begin{itemize}
\item[(i)]  Mean trunk inclination
\item[(ii)] Trunk angular excursion (i.e., net angular displacement)
\item[(iii)] Mean trunk angular velocity 
\item[(iv)] The phase of max. trunk flexion w.r.t. the stride cycle 
\end{itemize}

Experiments on human running show that for speeds of \SIrange[range-phrase=-,range-units=single]{2.5}{6}{\metre\per\second},  the mean trunk inclination increases from \SIrange[range-units=single]{8}{14}{\degree} \cite{Kunz_1981, Schache_1999},  along with a trunk angular excursion that extends from \SIrange[range-units=single]{2.5}{5}{\degree} \cite{Thorstensson_1984, Schache_1999} and a mean trunk angular velocity that increases from \SIrange[range-units=single]{5}{35}{\degree\per\second} \cite{Heitcamp_2012}. Meanwhile, the phase of maximum trunk flexion shifts from \SIrange[range-units=single]{30}{20}{\percent} of the strike cycle; so that the maximum trunk lean occurs closer to the aerial phase as the speed increases \cite{Heitcamp_2012}. These observations indicate that humans might utilize the trunk’s inertia and its oscillations to achieve their impressive locomotor energetics, speed, and robustness. Humanoid robots could similarly benefit from human-inspired trunk pitch strategies \cite{Williams_1987}.  
%
%

The spring-loaded inverted pendulum (SLIP) model captures the essential characteristics of running. It consists of a point-mass body attached to a massless, springy leg \cite{Blickhan_1989}. This model can be extended with a rigid trunk (\TSLIP), which is actuated by a torque at the hip \cite{Maus_1982, Sharbafi_2017}. A recent method to determine the hip torque is based on the \VP concept, where the ground reaction forces are redirected to intersect at a point \emph{above} the center of mass (\CoM). This phenomenon has been observed experimentally in human walking \cite{Maus_2010, Sharbafi_2017,Vielemeyer_2019}, running \cite{Maus_1982}, and other animal gaits \cite{Andrada_2014, Blickhan_2015}. Previous research often interprets the \VP as a \emph{pivot} point that provides a pendulum-like support and implements it as the target variable of control to achieve postural stability. In contrast to a \VP {\color{color_VPBlue}\text{ }above} the \CoM (\VPa), a recent study with parkinson patients provides evidence~to~a \VP {\color{color_VPRose}\text{ }below} \CoM (\VPb) in walking gaits \cite{Scholl_2018}. As an important novelty, here we focus our research on the effect of \VP location on trunk oscillations.

Most humanoid robots stabilize their trunk position at a fixed angle without any trunk movements \cite{Raibert_1986, Westervelt_2007}. An exception are the robots WL-12RV and Honda ASIMO, which create trunk accelerations to address unstable dynamic situations \cite{Shigemi_2019, Yamaguchi_1993}. The control strategy in ASIMO extends static walking and uses the trunk reaction to recover stability while running. Another approach is shown in the ATRIAS robot, which implements a \VP target \cite{Peekema_2015} or a sinusoidal reference trajectory \cite{Rezazadeh_2015} to create trunk movements for walking. While this approach works in practice, it is not well understood how the reference for the \VP affect the whole body motion.

In this work, we implement a \TSLIP model with different \VP targets. We then systematically compare the resulting gaits for the speeds of \SIrange[range-phrase=-,range-units=single]{4}{10}{\meter\per\second}, which is the physical limit for human running \cite{Bramble_2004}. We show that

\begin{itemize}
\item[(a)] VP above the CoM (\VPa) generates backward trunk motion that is observed in the single stance phase of human walking \cite{Aminiaghdam_2017, Thorstensson_1984}; whereas a VP below the~\CoM (\VPb) causes forward motion that is seen in running~\cite{Thorstensson_1984};
\item[(b)] \VPa results in smaller trunk angular excursions, as opposed to \VPb with the same \VP radius (i.e., the same distance between \VP and \CoM);
\item[(c)] the trunk angular excursion and mean trunk angular velocity increase with speed for the same VP radius. 
\end{itemize}

We demonstrate that the standard \TSLIP model is not capable of predicting leg length velocity accurately, and propose a bilinear damper to obtain smooth force profiles.  Finally, our findings support the hypothesis that trunk motions can be used in favor of energy economy.  We observe that a \VPb reduces the leg loading, which makes it appealing to place the \VP further down. However, a lower \VP shifts the power requirements to the hip, and increases peak hip torques.  Therefore there is a trade-off between leg loading and the peak torque when choosing the \VP location.

\section{SIMULATION MODEL} \label{sec:SimModel}
\begin{figure}[b!]
\centering
{\includegraphics[width=1\columnwidth] 
{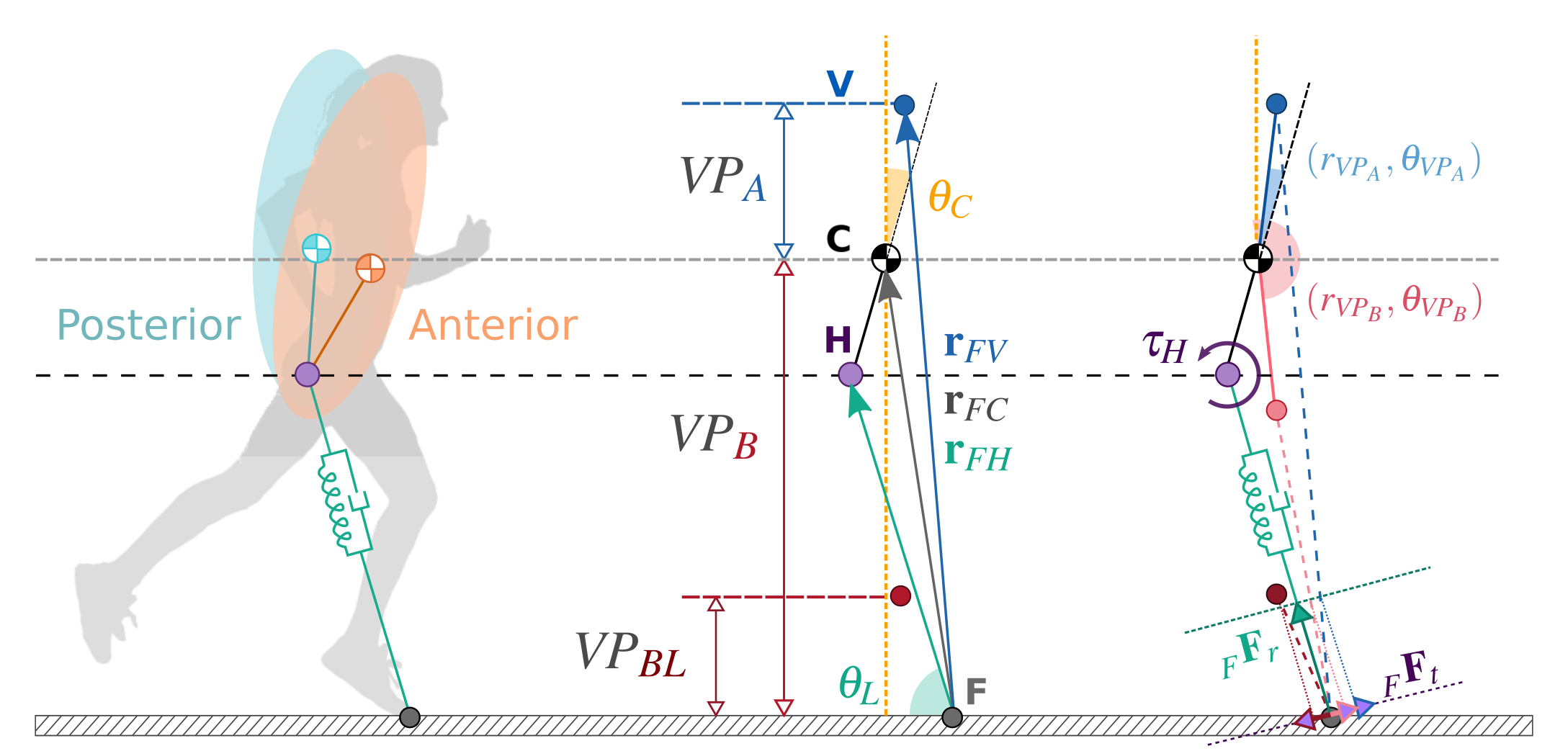}}
\captionof{figure}{\TSLIP model and vector definitions for the human morphology. {\protect \markerCW} region demonstrates clockwise trunk rotation, whereas {\protect \markerCCW} is the opposite rotation. The letters $\mathrm{V, \:  C, \: H}$ denote the virtual point, \CoM, hip respectively. The position vectors between points are referred as $\mathbf{r}_{FH}, \mathbf{r}_{FV}, \mathbf{r}_{FC}, \mathbf{r}_{FH}$. The angles $\mathbf{\theta}_{L}, \mathbf{\theta}_{C},\mathbf{\theta}_{VP}$ are the leg, trunk and \VP angles.}
 \label{fig:TSLIP}
\end{figure}

In this section, we describe our \TSLIP model applied in this work.
The \TSLIP model consists of a trunk with mass $m$ and moment of inertia $J$, which are attached to a massless leg at the hip. The leg is equipped with a parallel spring-damper mechanism (\cref{fig:TSLIP}). The system's hybrid dynamics are characterized by a flight phase, where the \CoM moves in a ballistic motion; and a stance phase, where the leg force and hip torque propel the body. The transition between these phases occurs at touch-down (TD) when the foot point comes in contact with the ground, and take-off (TO) when the \GRF becomes zero or the leg reaches its rest length $l_{0}$. 

The equations of the motion for the \CoM state $(x_{C},  y_{C},  \theta_{C} ) $ during the stance phase can be written as, 
\begin{equation}
m \begin{bmatrix}  \ddot{x}_{C} \\  \ddot{y}_{C} \end{bmatrix} =   \prescript{}{F}{\mathbf{F}}_{a} +  \prescript{}{F}{\mathbf{F}}_{t} + g \text{ and }
J\, \ddot{\theta}_{C} = \minus {\mathbf{r}}_{FC} \times  (  \prescript{}{F}{\mathbf{F}}_{a} +  \prescript{}{F}{\mathbf{F}}_{t}),
\label{eqn:EoM} 
\end{equation}
where the leg spring-damper forces $F_{sp}$ and $F_{dp}$ generate the axial component of the \GRF in foot frame $ \prescript{}{F}{\mathbf{F}}_{a}$, and the hip torque ${\tau}_{H}$  generates the tangential component $ \prescript{}{F}{\mathbf{F}}_{t}$,

\begin{equation}
\begin{aligned}
\prescript{}{F}{\mathbf{F}}_{a}  &= \left(F_{sp} \minus F_{dp} \right)  
 & {\begin{bmatrix*}[c] \minus \cos(\theta_{L}) &   \sin(\theta_{L})  \end{bmatrix*} }^\mathsf{T}, \\
\prescript{}{F}{\mathbf{F}}_{t}  &= \left( \scaleto{\sfrac{\minus \tau_{H}}{l_{L}}}{12pt} \right)  
& {\begin{bmatrix*}[c]  \sin(\theta_{L}) &   \minus \cos(\theta_{L})  \end{bmatrix*} }^\mathsf{T}.
 \label{eqn:FrFt} 
\end{aligned}
\end{equation}
%
We implemented a linear leg spring, with force $F_{sp} \text{=}k (l_{0}-l)$, with spring constant $k$, leg length $l$, and leg rest length $l_{0}$. The damping is often chosen to be linear; we explain our choice of nonlinear damping ($F_{dp}$) in \cref{sec:Modifications}.

Since the leg is passively compliant, trunk pitch motions can only be controlled via the hip torque $\tau_{H}$. We select $\tau_{H}$, such that the \GRF points to a \VP, which is characterized by a radius $r_{VP}$, and an angle $\theta_{VP}$ relative to the \CoM (\cref{fig:TSLIP}),
\begin{equation}
\begin{aligned}
\tau_{H} &= \tau_{VP} =  \prescript{}{F}{\mathbf{F}}_{a} \times  \left[   \frac{\mathbf{r}_{FV} \times \mathbf{r}_{FH} }{\mathbf{r}_{FV} \cdot \mathbf{r}_{FH}}\right]   \times  l,  \\
\mathbf{r}_{FV} &= \mathbf{r}_{FC}  + r_{VP}    \begin{bmatrix*}[r] \minus \sin \left( \theta_{C}+\theta_{VP} \right) \\ \cos \left( \theta_{C}+\theta_{VP} \right) \end{bmatrix*}.
\end{aligned}
\label{eqn:tauVP}
\end{equation}

Our model has seven morphological parameters, which are selected to match a human of \SI{80}{\kilogram}, with \SI{1}{\meter} leg (\cref{tab:ModelPrm}).

\begin{table}[h!]                                                                                                                                                                                                                                                                                                                                                                                                                                                                                                                                                                 
\centering                                                                                                                                                                                                                                                                                                                                                                                                                                                                                                                                                                          
\captionsetup{justification=centering}
\caption{Model parameters for TSLIP model}
\label{tab:ModelPrm}
\begin{adjustbox}{width=0.97\textwidth}
\begin{tabular}{@{} l| c c c c c  @{}}
\multicolumn{1}{l}{Name} & \multicolumn{1}{c}{Symbol} &  \multicolumn{1}{c}{Units} &   \multicolumn{1}{c}{Literature} & \multicolumn{1}{c}{Chosen} & \multicolumn{1}{c}{Reference}  \\
\hline
\hspace{1mm} mass & $\mathit{m}$  & \si{\kilogram}    &  60-80   & 80 &   {\cite{Sharbafi_2013, Sharbafi_2017}}  \\
\hspace{1mm} moment of inertia & $\mathit{J}$    & \si{\kilogram\meter\squared} & 5 & 5 &  {\cite{ Sharbafi_2013, deLeva_1996}}   \\ 
\hspace{1mm} leg stiffness & $\mathit{k}$  & \si{\kilo\newton\per\meter} &16-26 & 18 &  {\cite{Sharbafi_2013, McMahon_1990}}   \\
\hspace{1mm} leg length & $\mathit{l_{0}}$  & \si{\meter}   &  1  &  1 &  {\cite{Sharbafi_2013, Sharbafi_2017}}  \\
\hspace{1mm} leg angle at TD & $ \mathit{\theta_{L}^{TD}}$ & (\si{\degree}) & 78-71 & $\mathit{f_{H}(\dot{x})}$ & {\cite{Sharbafi_2013, McMahon_1990}}  \\
\hspace{1mm} dist. Hip-CoM & $\mathit{r_{HC}}$  & \si{\meter} & 0.1 &0.1 & {\cite{Sharbafi_2013, Wojtusch_2015}} 
\end{tabular}
\end{adjustbox}
\vspace{-4mm}
\end{table}

\section{MODIFICATIONS} \label{sec:Modifications}
In this section, we modify the leg damping to match the human data and introduce a novel control scheme that creates a range of feasible running gaits and trunk oscillations.

\subsubsection{Model Modifications}\label{subsec:ModelMod}
\begin{figure}[th!]
\centering
\includegraphics [width=\columnwidth]
   {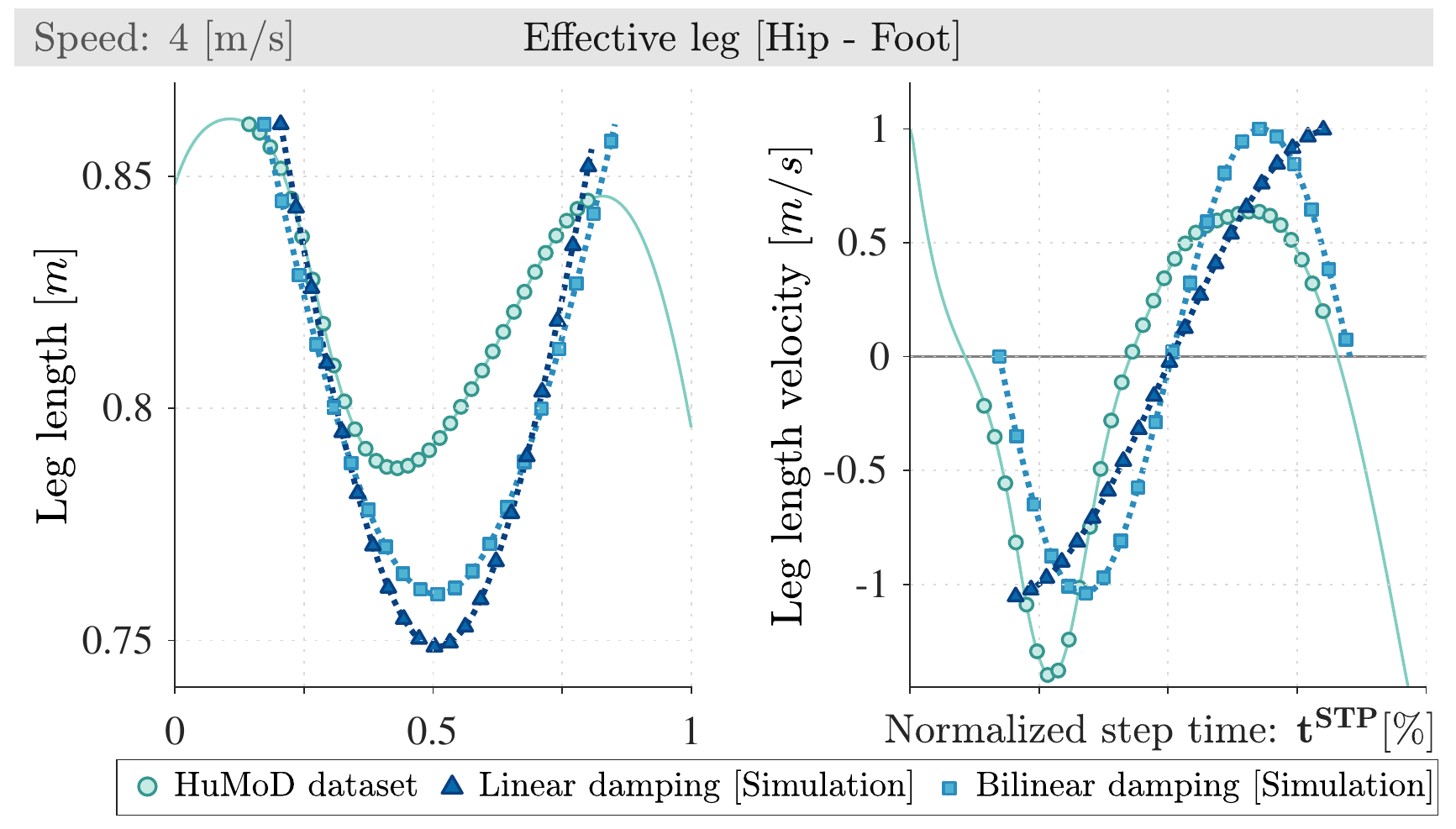}
\caption{A linear leg damper causes non-zero leg length velocities at TD/TO, which yield discontinuous damping forces. We introduce a bilinear damper to obtain more human-like behavior. The leg lengths are offset to the same TD value and velocities are normalized. The biomechanical data is estimated from the HuMoD dataset \cite{Wojtusch_2015} using \textit{de Leva} \cite{deLeva_1996} body segment parameters.}
\label{fig:BilinearDamping}
\end{figure}

In a \TSLIP simulation with no other energy sinks, linear leg dampers are often introduced to absorb the energy inserted by the hip torque \cite{Sharbafi_2013, Andrada_2014}.  However, this leads to non-zero leg length velocities at TD/TO, and thus discontinuous leg forces ({\protect \markerDmpLin}, \cref{fig:BilinearDamping}). In contrast, this is not observed in humans ({\protect \markerHuMoD}). To obtain realistic behavior, we use a bilinear damping with force
\begin{equation}
F_{dp} =  c \ \dot{l} \ (l_{0}-l) = c \ \dot{l} \ \Delta l ,
\label{eqn:BilinearDampingForce} 
\end{equation}
where $\dot{l}$ is the leg length velocity and $c$ is the damping~coefficient.  
This damping leads to smooth damping forces ({\protect \markerDmpBiLin})~and is mathematically similar to the hill-type muscle model \cite{Abraham_2015}.
%
%
\subsubsection{Control Strategy}\label{subsec:ControlMod}
To regulate the leg angle at TD, $\theta_{L}^{TD}$, we extend the method developed by \cite{Raibert_1986}. In particular, at the $i\,$th step, we choose 
\begin{equation*}
 \theta_{L}^{TD} \, |_{i} =  \theta_{L}^{TD} \, |_{i \mminus 1} +
 k_{\dot{x}0}    (  \Delta \dot{x}_{C}^{AP} \, |_{0}^{i} )
 +  k_{\dot{x}}       (  \Delta \dot{x}_{C}^{AP} \, |_{i  \mminus 1}^{i}  )
 + k_{y}                (\Delta {y}_{C}^{AP} \, |_{i \mminus 1}^{i}),
 \label{eqn:thetaL0}
\end{equation*}
where $k_{\dot{x}0}$, $k_{\dot{x}}$, and $k_{y}$ are the controller gains. Here, $\Delta \dot{x} |_{0}^{i}$ is the difference in apex velocity $\dot{x}$ between time steps $0$ and~$i$.

To generate trunk oscillations, we set the hip torque using a \VP as in \cref{eqn:tauVP}. However, this method is not robust~and is highly sensitive to initial conditions and parameter variations~\cite{Sharbafi_2013}. One way to overcome this is to modify $(r_{VP},\theta_{VP})$ at every time step using, for example, a linear quadratic regulator control \cite{Sharbafi_2013}. However, this controller can converge to any arbitrary $r_{VP}$ and is not suitable for our~purpose.

In this work, we want to investigate the impact of a fixed $r_{VP}$ on the magnitude and direction of trunk oscillations. To this end, we propose a new incremental control strategy that facilitates \VP gaits for a fixed $r_{VP}$ starting from various initial condition and for any model parameters.

\begin{figure}[t!]
\centering
  \includegraphics [width=\columnwidth] 
  {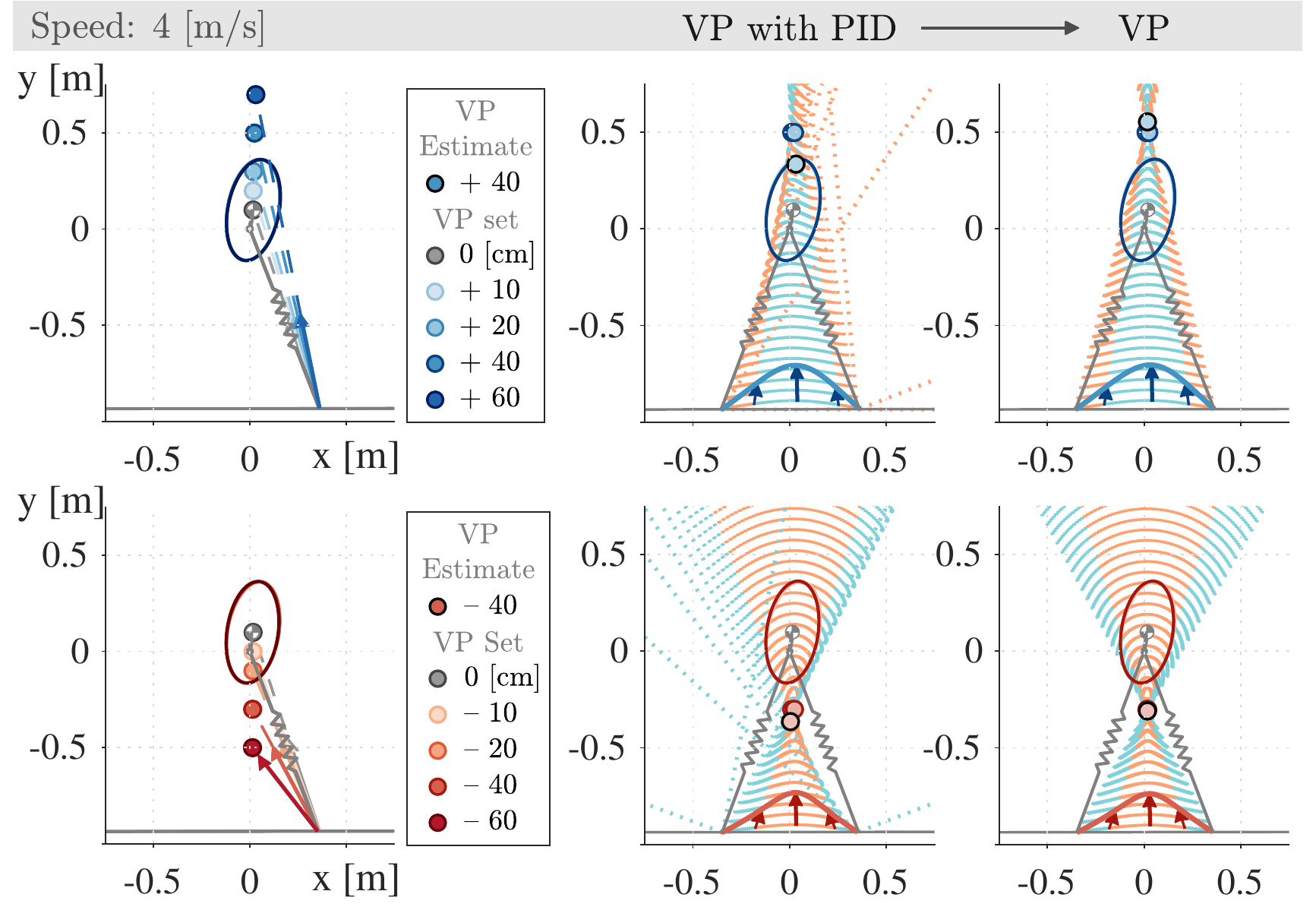}
\caption{Experimental setup (on left) and resulting \GRF patterns with Eq. \cref{eqn:tauH} and without \cref{eqn:vppLIN} PID reinforcement are plotted in non-rotating hip frame. The estimated \VP is marked with black rimmed circles. }
\label{fig:VP}
\end{figure}

First, we stabilize the \VP controller with an additive PID term, $\tau_{H} = \tau_{VP} +\tau_{PID}$, where
\begin{equation}
\tau_{PID} = k_{p}  \left( \theta^{Des}_{C} \minus \theta_{C} \right) 
+ k_{d} \left( \dot{\theta}^{Des}_{C} \minus \dot{\theta}_{C} \right)
+ k_{i}  \left( \theta_{C}^{Err} \right)  
\label{eqn:tauH} 
\end{equation}
with desired mean body pitch $\theta^{Des}_{C}$, pitch rate $\dot{\theta}^{Des}_{C}$, pitch integral error $\theta_{C}^{Err}$, and control gains $k_{p},k_{d},k_{i}$. Applying this controller to the \TSLIP model for two opposing positions of the \VP (above-below) leads to a diverted \GRF(\cref{fig:VP}, center column). The dotted lines represent the \GRF for clockwise ({\protect \markerCW}) and counterclockwise ({\protect \markerCCW}) trunk rotation. The magnitude of the \GRF is plotted with red/blue lines at the bottom. The VP+PID controller does not focus the \GRF at a single point as desired, but instead keeps the \GRF close to a fixed point.

After the VP+PID controller converges to steady-state motion, we impose the \GRF focus on a fixed point by disabling the PID, i.e., $ \tau_{PID} = 0$, and adapting the \VP angle,
\begin{equation}
\theta_{VP} \ |_{i} =\theta_{VP} \ |_{i-1} + k_{vp} \left( \theta_{C}^{Des}- \Delta\theta_{C}   \right ).
\label{eqn:vppLIN} 
\end{equation}
This controller adjusts the \VP angle based on the difference between the desired mean body angle $ \theta_{C}^{Des}$ and the observed one over the last step, $\Delta\theta_{C}$. It converges to a fixed \VP solution, where we can compare the pitch oscillation characteristics (\cref{fig:VP}, right column). 

\section{SIMULATION RESULTS} \label{sec:SimuationResults}
In this section,  we describe our simulation setup to compare trunk oscillations and analyze our results.

\subsection{Simulation Setup}\label{subsec:ExpSetup}
In our simulations, we want to evaluate trunk oscillations up to at least \SI{5}{\degree}, as observed in humans. For this reason, we sweep \VP targets over  $r_{VP} = \pm [0,\, 20,\, 40,\, 60]\,$\si{\centi\meter} and keep the desired mean body pitch angle fixed to  \SI{10}{\degree} (\cref{fig:VP}, left column). We define the target \VP angle w.r.t. body coordinates for \VPa and w.r.t. world coordinates for \VPb.

Besides the \VP location, the choice of control and damping parameters influence the flow of the motion. In biomechanics, gaits can be characterized by the duty factor (DF). DF is the ratio of the leg contact time to the stride period and is correlated to the magnitude of the \GRF \cite{Bishop_2018}.
%
\begin{figure}[b!]
\centering
  \includegraphics [width=0.9\columnwidth]
  {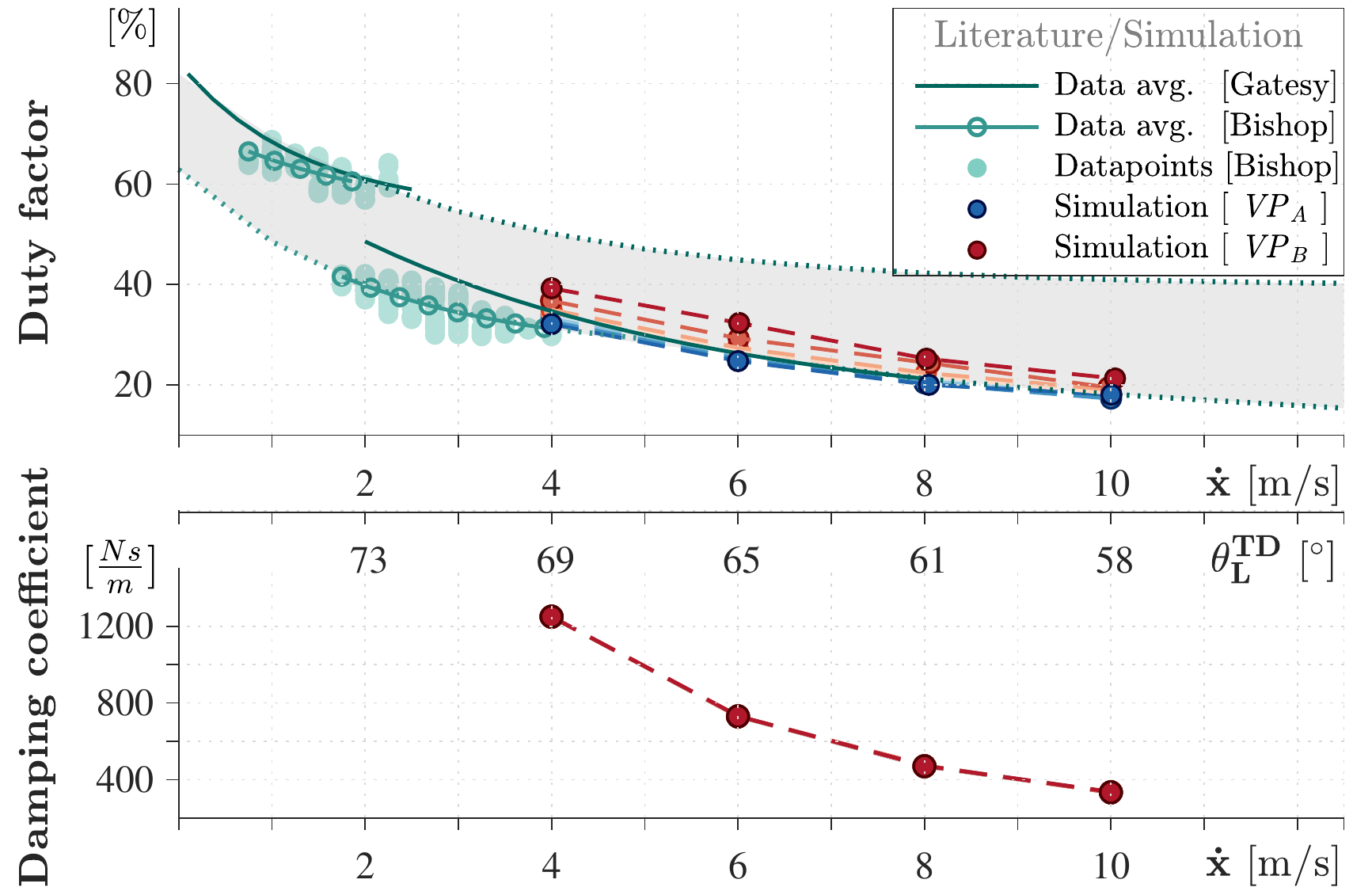}
\caption{Control parameters and damping are adjusted so that the simulation's duty factor lie in the grey shaded region estimated from biomechanical experiments \cite{Gatesy_1991, Bishop_2018}.  }
\label{fig:Lit_duty}
\end{figure}
We tune control gains to match DFs of simulated gaits to the data in \cite{Gatesy_1991, Bishop_2018}. Consequently, the DF, leg angle at TD, and damping coefficient decrease with speed (\cref{fig:Lit_duty}). We apply the same damping coefficient for all \VP targets of equal speed.

\subsection{Analysis of the Ground Reaction Force (\GRF)}\label{subsec:GRF}
\subsubsection{Effect of Damping}\label{subsubsec:Effect_c}
We investigate how the damping in our model affects the system dynamics. We subtract the component of \GRF produced by damping (\cref{fig:c_tau_GRF}, light-colored lines) and detect a phase shift to left when damping is present (\cref{fig:c_tau_GRF}, dark-colored lines, horizontal arrows).

The area under the horizontal GRF (\GRFx) represents~the fore-aft impulse, which is proportional to the forward acc/deceleration. Damping contributes to braking during the TD-MS phase (MS is the midstance) and hinders accelerating over the MS-TO phase. The damping contribution can be seen with the decrement in \GRFx magnitude, which is denoted with vertical arrows (\cref{fig:c_tau_GRF}, right column).
\newline 

\subsubsection{Effect of VP Location}\label{subsubsec:Effect_VPloc} 
\begin{figure}[b!]
\centering
 \includegraphics [width=\columnwidth] 
  {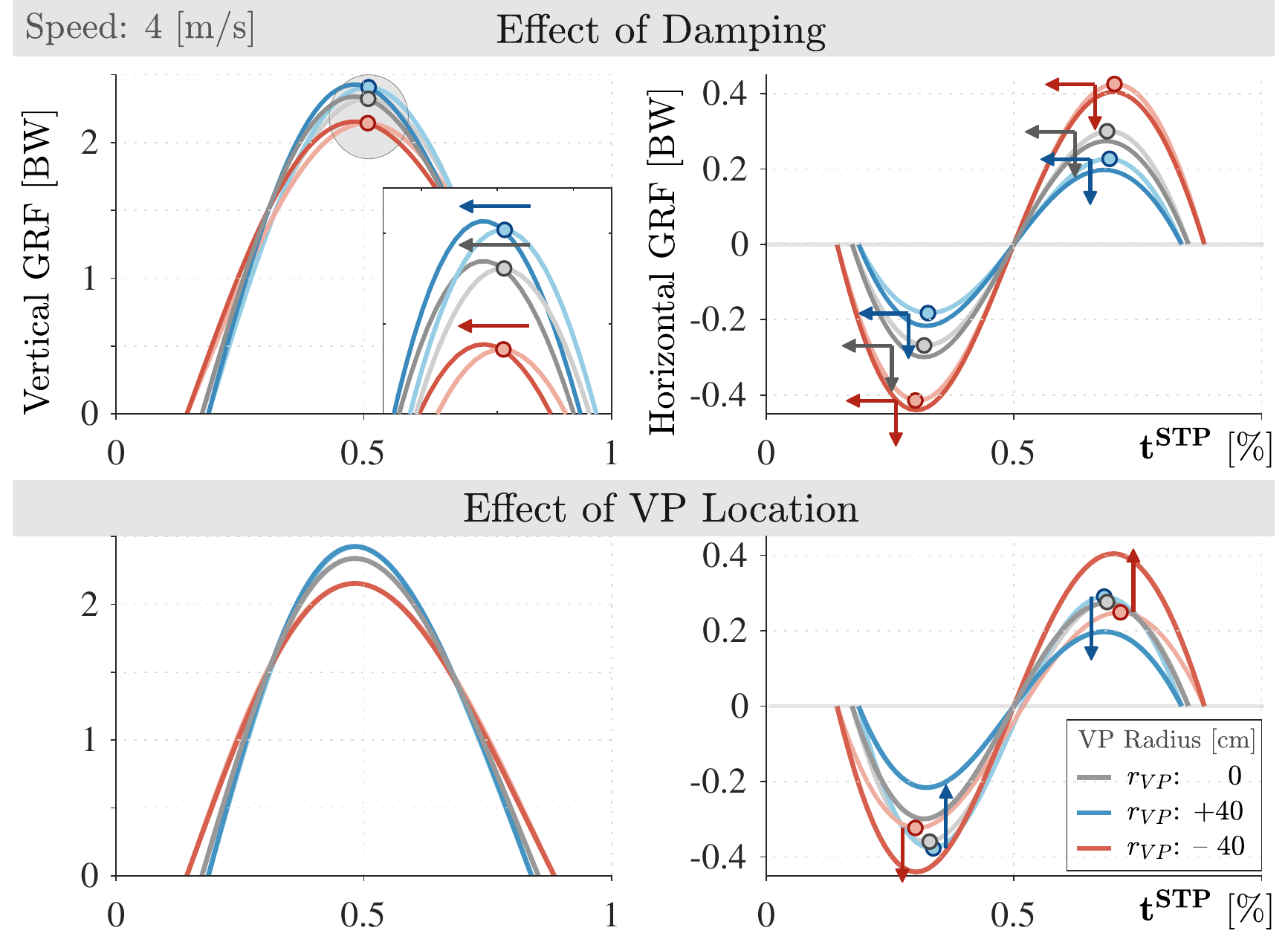}
\caption{The light-colored lines represent the \GRF, when the component created by damping (top) and hip torque (bottom) are subtracted. We observe a temporal shift in the GRF with the addition of damping. Damping contributes to decelerating over TD-MS and accelerating over MS-TO by changing the \GRFx magnitude.}
\label{fig:c_tau_GRF}
\end{figure}

We compare three \VP locations: \VP at the \CoM, above and below it. The \VP location affects the decomposition of the \GRF vector, and implicitly determines the cooperation between the leg force and force generated by the hip torque. When \VPb, leg and hip forces contribute to the \GRFx synergistically and \GRFy antagonistically, which is illustrated with blue/green arrows (\cref{fig:TSLIP}, right column). The \GRFy is created mainly by the leg force, and the contribution of hip is negligible (\cref{fig:c_tau_GRF}, bottom left). On the other hand, the forces constructing \GRFx have a similar order of magnitude. \VPb makes the hip force act along the same direction as the leg, and hence decreases the magnitude of \GRFx, as indicated with the red vertical arrows \cref{fig:c_tau_GRF}, bottom right). In other words,  \VPb causes the system to accelerate and decelerate more in the fore-aft direction. The \VPa has the inverse effect and causes smaller accelerations/decelerations.

\subsection{Trunk Pitch Oscillations}\label{subsec:TrunkPitchOscillations}
\begin{figure}[b!]
\centering
  \includegraphics [width=\columnwidth] 
  {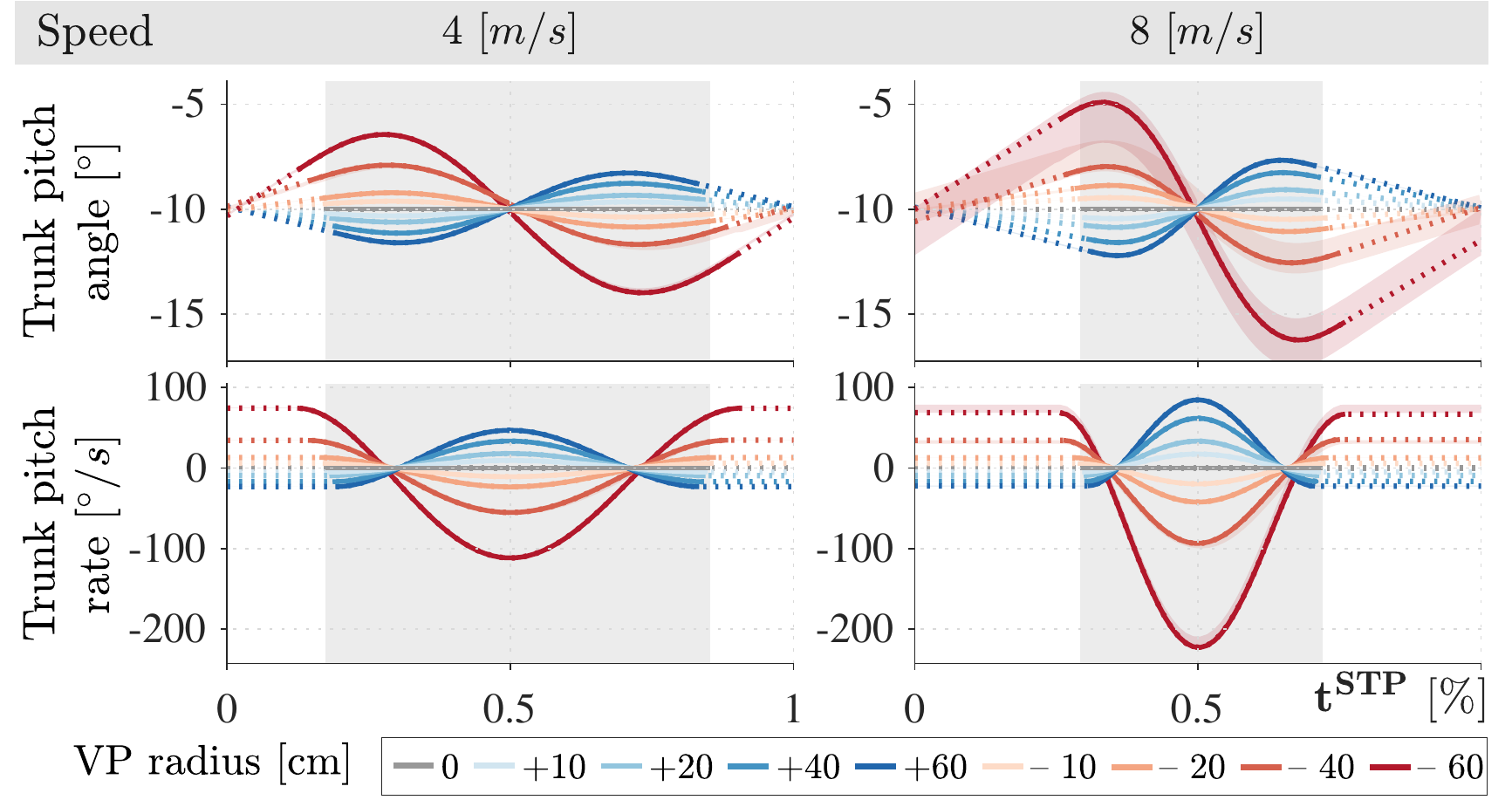}
\caption{\VPa produces backward and \VPb generates forward trunk pitching. The magnitude and rate of pitch oscillations increase with the \VP radius.  }
\label{fig:thB}
\end{figure}

Our parameter sweep over $r_{VP}$ reveals that a \VPa produces backward trunk pitching during the stance phase (\cref{fig:thB}, blue lines). A \VPb leads to forward trunk motion (red lines), which contradicts with \cite{Maus_1982, Maus_2010} that estimates a \VPa using the \GRF measurements of human walking and running. 
We propose two potential explanations for this discrepancy. First, the trunk pitching motion is measured \SI{180}{\degree} out of~phase with the whole body pitching in human walking \cite{Mueller_2017}. It hints that the model with a \VPa might predict the whole body motions instead of the trunk. Second, the \GRF data is often cropped at TD-TO due to undesired effects of impact and ankle push-off \cite{Maus_1982}, which might corrupt the estimation of~\VP.

\begin{figure}[b!]
\centering
  \includegraphics [width=\columnwidth] 
  {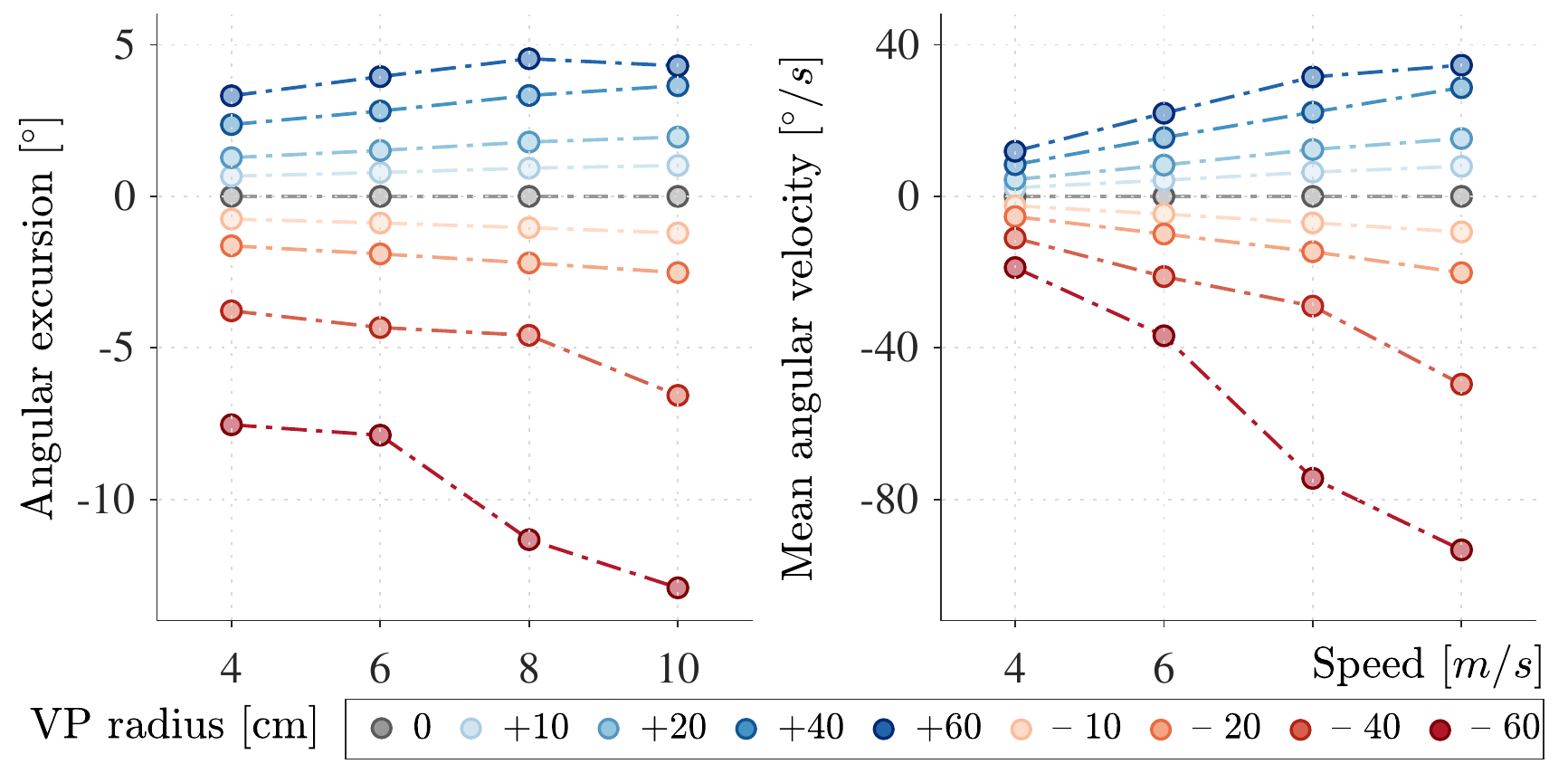}
\caption{Trunk angular excursion and peak angular rate increase with speed and increasing absolute \VP radius.}
\label{fig:thBdthB_dx}
\end{figure}

The trunk oscillates with a larger amplitude for \VPb compared to \VPa, given an equal \VP radius. The larger amplitude is caused by the larger moment arm of the \GRF around the \CoM with \VPb.  In accordance, at running speeds between \SIrange[range-phrase=-,range-units=single]{4}{10}{m/s} (\cref{fig:thBdthB_dx}), the trunk angular excursion increases up to \SI{5}{\degree} for \VPa and \SI{14}{\degree} for \VPb. The mean trunk angular velocity increases up to \SI{38}{\degree\per\second} for \VPa and $\minus$\SI{85}{\degree\per\second} for \VPb.

The trunk oscillates more at higher speeds, albeit an equal \VP target (\cref{fig:thBdthB_dx}). As the speed increases (\cref{tab:dxProgression}), we see that the leg force $\prescript{}{F}{{F}}_{a}$ increases despite the rise in damping $F_{dp}$, since $F_{sp}$ is significantly larger than $F_{dp}$. The \VP method imposes a linear relation between the leg and hip forces. Therefore the \GRF gets larger over speed. Higher \GRF creates a larger angular moment around the \CoM, which {results in a larger trunk oscillation.
\begin{table}[t!]                                                                                                                                                                                                                                                                                                                                                                                                                                                                                                                                                                 
\centering
\caption{Evolution of the leg forces w.r.t. the increase in forward speed.}
\label{tab:dxProgression}
\begin{adjustbox}{width=0.6\columnwidth}
\begin{tabular}{ l| c c c c c c }
\multicolumn{1}{l}{$\dot{x}\ \scaleto{\left[ \,\text{\si{\meter\per\second}} \, \right]}{10 pt}$} & \multicolumn{1}{c}{$c$} &  \multicolumn{1}{c}{$\Delta l$} &   \multicolumn{1}{c}{$\dot{l}$} & \multicolumn{1}{c}{$F_{sp}$} & \multicolumn{1}{c}{$F_{dp}$}  &   \multicolumn{1}{c}{$\prescript{}{F}{{F}}_{a}$} \\
\hline
\hspace{2mm} 4 & {$3\: c$} & {$\Delta l$} & {$\dot{l}$} & {$15\: F$} & {$F$} & {$14\: F$}  \\
\hspace{1mm} 10 &   {$\: c$} &  {$2\: \Delta l$} & {$6\: \dot{l}$}  & {$30\: F$}  & {$4\: F$}  & {$26\: F$} 
\end{tabular}
\end{adjustbox}
\vspace{-4mm}
\end{table}

\subsection{Energy Considerations}\label{subsec:Energy}
In this section, we investigate the energetic consequences our \VP target locations. Simulated gaits converge to a periodic steady-state solution, where the system's energy is conserved over steps. However, the energy fluctuates within a single step. The hip torque adds energy to propel to body, and the leg damper depletes this excess energy to prevent the body from accelerating or decelerating. Effectively, the work done by the leg (g) and the hip (h) have the same magnitude and opposite sign (\cref{fig:WorkDistribution}). 

In the course of a single step, the hip torque generated by the \VPa injects energy to the system in the initial stages of the stance phase (i, {\protect\arrowWorkP\,}), and removes some energy towards the end (i, {\protect\arrowWorkN\,}, \cref{fig:Kinetics}). The switch occurs after MS ({\protect \markerVPa}), and the hip  energy end up in a net positive value at the end of the stance phase ({\protect \markerVPa}). For the \VPb with a radius smaller than $\mminus$ \SI{30}{\centi\meter}, this relation holds as well. When the \VP radius gets larger than \SI{30}{\centi\meter} (\VPbl), the \VP becomes located below the leg at TD, which changes the sign of the hip torque. Consequently, the hip depletes energy first, reaches to a maximum  before MS ({\protect \markerVPb}), and injects a greater energy after that ({\,\protect\arrowWorkP\,}) to reach to a net positive value ({\protect \markerVPbTri}). The energy min/maximum gets closer to the MS when $r_{VP}$ increases.

\begin{figure}[b!]
\centering
  \includegraphics [width=\columnwidth] 
  {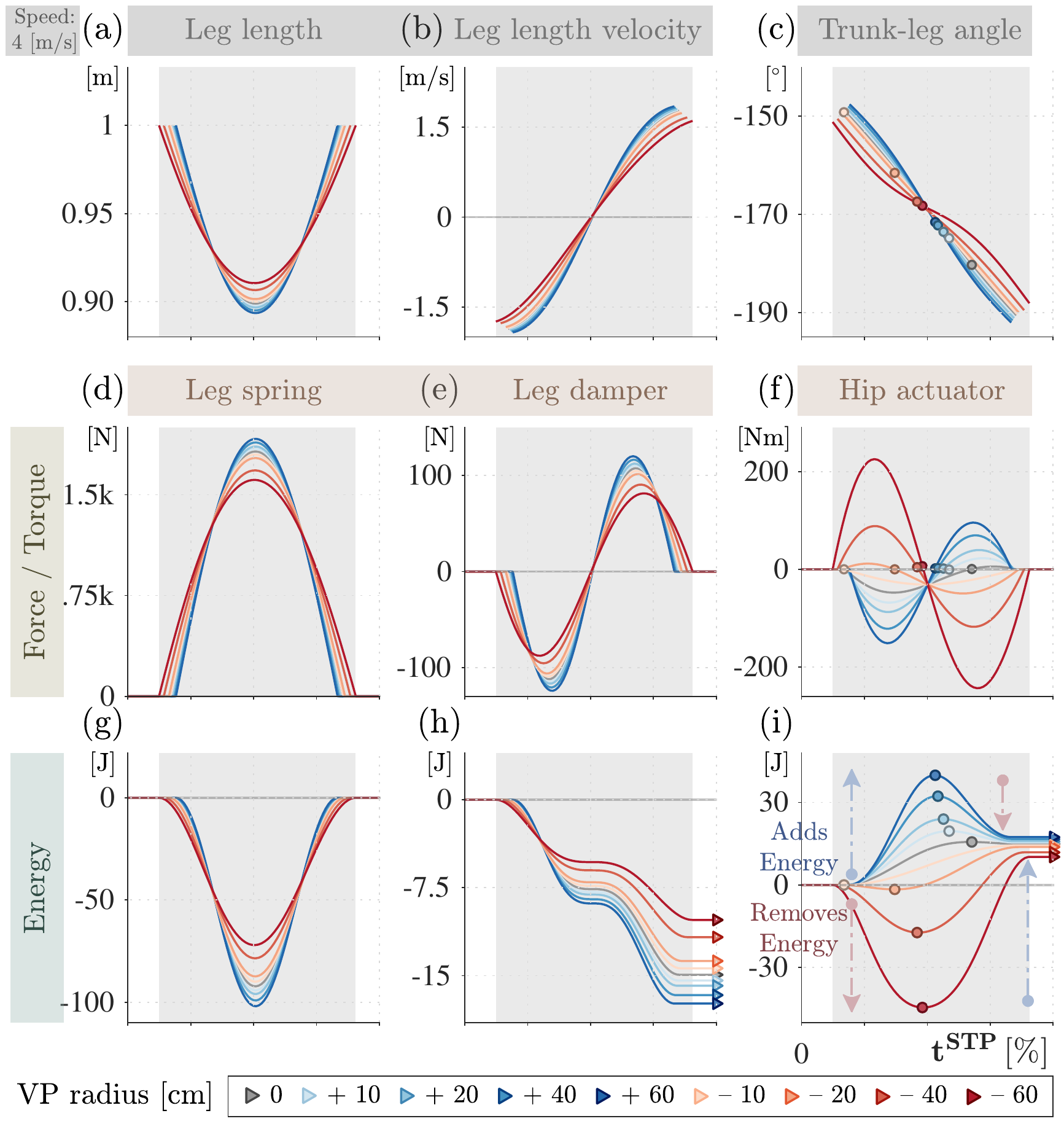}
\caption{The hip torque generated by \VPa injects and removes energy from the system, whereas for \VPbl the order reverses. Both methods yield a net positive hip work (marked with ({\protect \markerVPaTri},{\protect \markerVPbTri}), which leg damper has to remove. The \VP radius effects the phase of the energy reversal (marked with ({\protect \markerVPa},{\protect \markerVPb})}
\label{fig:Kinetics}
\end{figure}

\VPa yields higher net hip work (i,{\protect \markerVPaTri}\!) than \VPb, which~in return requires higher leg damping (h,{\protect \markerVPaTri}\!) to sustain the steady- state motion (\cref{fig:Kinetics}). This higher damping is provided by a larger $\Delta l \,$(a) and $\dot{l} \,$(b), given that the damping constant $c$ is kept constant for all \VP targets at the same speed. The higher $\Delta l$ yields higher spring forces (d) and energies (g) for~\VPa.

Our results show that the direction and magnitude of trunk oscillations determine the work distribution between the leg and hip. To analyze these two key factors, (1) we compare \VPa to \VPb for the same \VP radius and (2) we vary the \VP radius and analyze its effect for \VPa and \VPb separately.

\indent  \textrm{1.} {\color{color_BlueGrey}{Oscillation direction}} [\VPa$\,$vs.$\,$\VPb]: \VPa requires larger leg work (a,d,g) and higher net hip work (h, \cref{fig:WorkDistribution}). \\
\indent \textrm{2.}  {\color{color_BlueGrey}{Oscillation magnitude}} [var. $r_{VP}$]: \VPa with a larger radius requires higher leg (a,d,g) and net hip work (h), whereas its the opposite for \VPb (\cref{fig:WorkDistribution}). On the other hand, the magnitude of the positive and negative hip work depends on the \VP radius and increases with it (b,e). Consequently, the amount of positive and negative work performed by the hip becomes closer to that of the leg (c,f).

\begin{figure}[b!]
\centering
  \includegraphics [width=\columnwidth] 
  {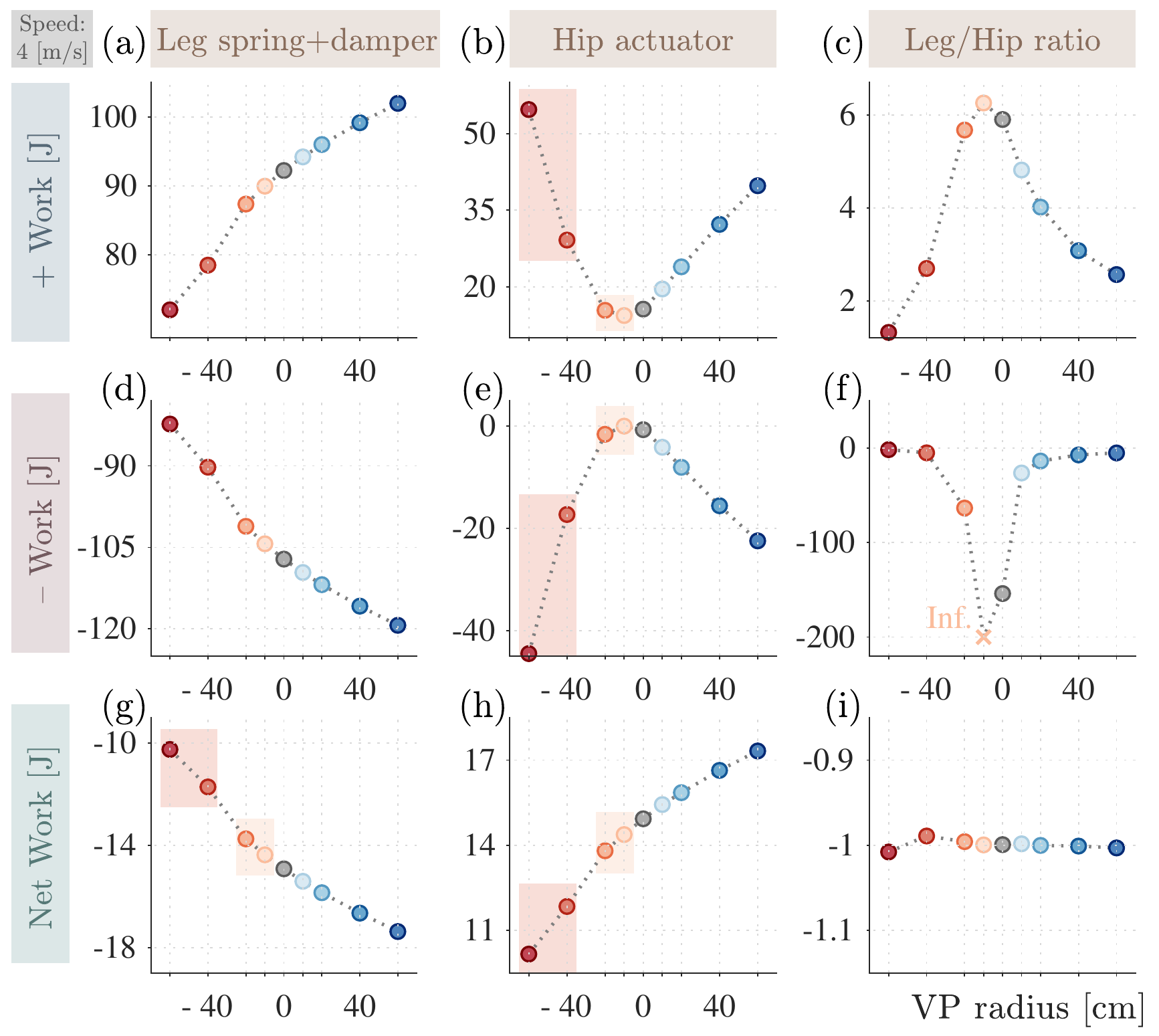}
\caption{Figure shows the positive, negative and the net work done by the leg and the hip. \VPb requires less the leg work and net hip work as opposed to \VPa. However, as \VPb is placed further down (\protect \markerVPbB), the absolute positive and negative hip energy requirements increase as well. }
\label{fig:WorkDistribution}
\end{figure}

In summary, a \VP below the \CoM in our running model reduces the leg loading and net hip work (\cref{fig:WorkDistribution}, shaded \protect \markerVPbA, \markerVPbB). As the \VP is placed further down (\protect \markerVPbB), the moment arm of the \GRF increases and higher hip torques become necessary to counteract the larger angular moments created. Intuitively, in human running, the \GRF forces the trunk to collapse forward at foot contact due to the posterior placement of the hip. If we place the \VPb, the trunk motion conforms to the \textit{natural pitching dynamics} of the trunk-leg system, which is to rotate forward, and assists the leg to decelerate in fore-aft ($x$) direction. After MS, the leg dynamics alone would force the trunk to extend when the trunk-leg angle decreases below $180^{\circ}$. \VPb imposes this transition to happen earlier. Meanwhile, it supports the leg in accelerating in $x$. The synergetic work of the hip and leg in $x$ leads to a reduced leg loading. However, as we place the \VP further down (\protect \markerVPbB), the increase in the moment arm of the \GRF (i.e., increase in angular momentum) dominates and and higher torques are needed to maintain postural stability. Importantly, our results corroborate previous findings in  \cite{Teng_2013}, who argue that the trunk posture influences lower extremity biomechanics and energetics by changing the load distribution in lower~limbs. 

\section*{CONCLUSION}\label{sec:Conclusion}
In this work, we investigated how the trunk pitch motion affects the dynamics and energetics of running. We implemented a TSLIP model with a virtual point (\VP) either above or below the center of mass (\CoM) to generate trunk oscillations. We showed that

\begin{itemize}
\item[(a)] \VP above the \CoM (\VPa) produces backward, below the \CoM (\VPb) produces forward trunk pitch motion,
\item[(b)]  \VPa creates smaller trunk angular excursions than \VPb for the equal \VP radius,
\item[(c)] the trunk angular excursion and mean trunk angular velocity increase with speed for the same \VP radius.
\end{itemize}

We also analyzed how the \VP location modified the work distribution between the hip and leg, and derived two key strategies to leverage the trunk oscillations in favor~of energetics. If the control goal is to minimize the leg loading, we suggest a \VPb with large radius (\VPbl), where the hip works synergistically with leg, at the expense of high peak hip torques. If we aim to minimize both the hip and leg work, we suggest \VPb with a radius smaller than \SI{30}{\centi\meter}. Our results can be extended to different bipedal morphologies with different parameterizations, and have potential implications for humanoid robot control. For instance, one can induce trunk motions and redistribute the joint work to counter motor torque limits or achieve a larger range of motion.

 \bibliographystyle{IEEEtran}
 \bibliography{root.bib}

\begin{thebibliography}{10}
\providecommand{\url}[1]{#1}
\csname url@rmstyle\endcsname
\providecommand{\newblock}{\relax}
\providecommand{\bibinfo}[2]{#2}
\providecommand\BIBentrySTDinterwordspacing{\spaceskip=0pt\relax}
\providecommand\BIBentryALTinterwordstretchfactor{4}
\providecommand\BIBentryALTinterwordspacing{\spaceskip=\fontdimen2\font plus
\BIBentryALTinterwordstretchfactor\fontdimen3\font minus
  \fontdimen4\font\relax}
\providecommand\BIBforeignlanguage[2]{{%
\expandafter\ifx\csname l@#1\endcsname\relax
\typeout{** WARNING: IEEEtran.bst: No hyphenation pattern has been}%
\typeout{** loaded for the language `#1'. Using the pattern for}%
\typeout{** the default language instead.}%
\else
\language=\csname l@#1\endcsname
\fi
#2}}

\bibitem{Raibert_1986}
M.~Raibert, ``{Hopping on One Leg in the Plane},'' in \emph{Legged Robots That
  Balance}.\hskip 1em plus 0.5em minus 0.4em\relax Cambridge, Massachusetts:
  The MIT Press, 1986, ch.~2, pp. 29--56.

\bibitem{Westervelt_2007}
E.~R. Westervelt, J.~W. Grizzle, C.~Chevallereau, J.~H. Choi, and B.~Morris,
  ``{Systematic Design of Within-Stride Feedback Controllers for Walking},'' in
  \emph{Feedback Control of Dynamic Bipedal Robot Locomotion}.\hskip 1em plus
  0.5em minus 0.4em\relax CRC Press, 2007, ch. 6-9.

\bibitem{Aminiaghdam_2017}
S.~Aminiaghdam, C.~Rode, R.~M{\"{u}}ller, and R.~Blickhan, ``{Increasing trunk
  flexion transforms human leg function into that of birds despite different
  leg morphology},'' \emph{Journal of Experimental Biology}, vol. 220, no.~3,
  pp. 478--486, 2017.

\bibitem{Heitcamp_2012}
L.~Heitkamp, ``{The Role of the Gluteus Maximus on Trunk Stability in Human
  Endurance Running},'' Ph.D. dissertation, University of Cincinatti, 2012.

\bibitem{Hinrichs_1987}
R.~N. Hinrichs, ``{Upper Extremity Function in Running. II: Angular Momentum
  Considerations},'' \emph{International Journal of Sport Biomechanics},
  vol.~3, no.~3, pp. 242--263, 1987.

\bibitem{Schache_1999}
A.~G. Schache, K.~L. Bennell, P.~D. Blanch, and T.~V. Wrigley, ``{The
  coordinated movement of the lumbo pelvic hip complex during running: a
  literature review},'' \emph{Gait {\&} Posture}, vol.~10, no.~1, pp. 30--47,
  1999.

\bibitem{Thorstensson_1984}
A.~Thorstensson, J.~Nilsson, H.~Carlson, and M.~R. Zomlefer, ``{Trunk movements
  in human locomotion},'' \emph{Acta Physiologica Scandinavica}, vol. 121,
  no.~1, pp. 9--22, 1984.

\bibitem{deLeva_1996}
P.~de~Leva, ``{Adjustments to Zatsiorsky-Seluyanov's segment inertia
  parameters},'' \emph{Journal of Biomechanics}, vol.~29, no.~9, pp.
  1223--1230, 1996.

\bibitem{Bramble_2004}
D.~M. Bramble and D.~E. Lieberman, ``{Endurance running and the evolution of
  Homo},'' \emph{Nature}, vol. 432, no. 7015, pp. 345--352, 2004.

\bibitem{Kunz_1981}
H.~Kunz and D.~A. Kaufmann, ``{Biomechanical analysis of sprinting: decathletes
  versus champions},'' \emph{British journal of sports medicine}, vol.~15,
  no.~3, pp. 177--181, 1981.

\bibitem{Williams_1987}
K.~R. Williams and P.~R. Cavanagh, ``{Relationship between distance running
  mechanics, running economy, and performance},'' \emph{Journal of Applied
  Physiology}, vol.~63, no.~3, pp. 1236--1245, 1987.

\bibitem{Blickhan_1989}
R.~Blickhan, ``{The spring-mass model for running and hopping},'' \emph{Journal
  of Biomechanics}, vol.~22, no.~11, pp. 1217--1227, 1989.

\bibitem{Maus_1982}
M.~H. Maus, ``{Stabilisierung des Oberk{\"{o}}rpers beim Rennen und Gehen},''
  Ph.D. dissertation, Friedrich Schiller University Jena, 1982.

\bibitem{Sharbafi_2017}
M.~A. Sharbafi, ``{Bioinspired template-based control of legged locomotion},''
  Ph.D. dissertation, Technical University of Darmstadt, 2017.

\bibitem{Maus_2010}
M.~H. Maus, S.~W. Lipfert, M.~Gross, J.~Rummel, and A.~Seyfarth, ``{Upright
  human gait did not provide a major mechanical challenge for our ancestors},''
  \emph{Nature Communications}, vol.~1, p.~70, 2010.

\bibitem{Vielemeyer_2019}
J.~Vielemeyer, E.~Grie{\ss}bach, and R.~M{\"{u}}ller, ``{Ground reaction forces
  intersect above the center of mass even when walking down visible and
  camouflaged curbs},'' \emph{The Journal of Experimental Biology}, pp.
  204--305, 2019.

\bibitem{Andrada_2014}
E.~Andrada, C.~Rode, Y.~Sutedja, J.~A. Nyakatura, and R.~Blickhan, ``{Trunk
  orientation causes asymmetries in leg function in small bird terrestrial
  locomotion},'' \emph{Proceedings of the Royal Society B: Biological
  Sciences}, vol. 281, no. 1797, 2014.

\bibitem{Blickhan_2015}
R.~Blickhan, E.~Andrada, R.~M{\"{u}}ller, C.~Rode, and N.~Ogihara,
  ``{Positioning the hip with respect to the COM: Consequences for leg
  operation},'' \emph{Journal of Theoretical Biology}, vol. 382, pp. 187--197,
  2015.

\bibitem{Scholl_2018}
\BIBentryALTinterwordspacing
P.~Scholl, ``{Modeling Postural Control in Parkinson's Disease},'' Ph.D.
  dissertation, Technical University of Darmstadt, 2018. [Online]. Available:
  \url{http://wiki.ifs-tud.de/abschlussarbeiten/msc/2018{\_}scholl}
\BIBentrySTDinterwordspacing

\bibitem{Shigemi_2019}
S.~Shigemi, ``{ASIMO and Humanoid Robot Research at Honda},'' in \emph{Humanoid
  Robotics: A Reference}, A.~Goswami and P.~Vadakkepat, Eds.\hskip 1em plus
  0.5em minus 0.4em\relax Springer, Dordrecht, 2019, pp. 55--90.

\bibitem{Yamaguchi_1993}
J.~I. Yamaguchi, A.~Takanishi, and I.~Kato, ``{Development of a biped walking
  robot compensating for three-axis moment by trunk motion},'' pp. 561--566,
  1993.

\bibitem{Peekema_2015}
A.~T. Peekema, ``{Template-Based Control of the Bipedal Robot ATRIAS},'' Ph.D.
  dissertation, Oregon State University, 2015.

\bibitem{Rezazadeh_2015}
S.~Rezazadeh and J.~W. Hurst, ``{Toward step-by-step synthesis of stable gaits
  for underactuated compliant legged robots},'' in \emph{IEEE International
  Conference on Robotics and Automation (ICRA)}, 2015, pp. 4532--4538.

\bibitem{Sharbafi_2013}
M.~A. Sharbafi, C.~Maufroy, M.~N. Ahmadabadi, M.~J. Yazdanpanah, and
  A.~Seyfarth, ``{Robust hopping based on virtual pendulum posture control},''
  \emph{Bioinspiration {\&} Biomimetics}, vol.~8, no.~3, 2013.

\bibitem{McMahon_1990}
T.~A. McMahon and G.~C. Cheng, ``{The mechanics of running: How does stiffness
  couple with speed?}'' \emph{Journal of Biomechanics}, vol.~23, pp. 65--78,
  1990.

\bibitem{Wojtusch_2015}
J.~Wojtusch and O.~von Stryk, ``{HuMoD - A Versatile and Open Database for the
  Investigation, Modeling and Simulation of Human Motion Dynamics on Actuation
  Level},'' in \emph{Proceedings of the IEEE-RAS International Conference on
  Humanoid Robots}, 2015, pp. 74--79.

\bibitem{Abraham_2015}
I.~Abraham, Z.~Shen, and J.~Seipel, ``{A Nonlinear Leg Damping Model for the
  Prediction of Running Forces and Stability},'' \emph{Journal of Computational
  and Nonlinear Dynamics}, vol.~10, no.~5, 2015.

\bibitem{Bishop_2018}
P.~J. Bishop, D.~F. Graham, L.~P. Lamas, J.~R. Hutchinson, J.~Rubenson, J.~A.
  Hancock, R.~S. Wilson, S.~A. Hocknull, R.~S. Barrett, D.~G. Lloyd, and C.~J.
  Clemente, ``{The influence of speed and size on avian terrestrial locomotor
  biomechanics: Predicting locomotion in extinct theropod dinosaurs},''
  \emph{PLOS ONE}, vol.~13, no.~2, 2018.

\bibitem{Gatesy_1991}
S.~M. Gatesy and A.~A. Biewener, ``{Bipedal locomotion: effects of speed, size
  and limb posture in birds and humans},'' \emph{Journal of Zoology}, vol. 224,
  no.~1, pp. 127--147, 1991.

\bibitem{Mueller_2017}
R.~M{\"{u}}ller, C.~Rode, S.~Aminiaghdam, J.~Vielemeyer, and R.~Blickhan,
  ``{Force direction patterns promote whole body stability even in hip-flexed
  walking, but not upper body stability in human upright walking},''
  \emph{Proceedings. Mathematical, physical, and engineering sciences}, vol.
  473, no. 2207, 2017.

\bibitem{Teng_2013}
H.-L. Teng and C.~M. Powers, ``{Influence of Sagittal Plane Trunk Posture on
  Lower Extermity Biomechanics During Running},'' Ph.D. dissertation,
  University of Southern California, 2013.

\end{thebibliography}

\end{document}